\documentclass[conference]{IEEEtran}
\IEEEoverridecommandlockouts
\usepackage{cite}
\usepackage{amsmath,amssymb,amsfonts}
\usepackage{algorithmic}
\usepackage{graphicx}
\usepackage{textcomp}
\usepackage{xcolor}
\usepackage{tabularx}
\def\BibTeX{{\rm B\kern-.05em{\sc i\kern-.025em b}\kern-.08em
    T\kern-.1667em\lower.7ex\hbox{E}\kern-.125emX}}
\begin{document}

\title{Predicting Three Types of Freezing of Gait Events Using Deep Learning Models\\
}

\author{\IEEEauthorblockN{1\textsuperscript{st} Wen Tao Mo}
\IEEEauthorblockA{\textit{Engineering Science} \\
\textit{University of Toronto}\\
Toronto, Canada \\
m.mo@mail.utoronto.ca}
\and
\IEEEauthorblockN{2\textsuperscript{nd} Jonathan H. Chan}
\IEEEauthorblockA{\textit{Innovative Cognitive Computing (IC2) Research Center, School of Information Technology} \\
\textit{King Mongkut's University of Technology Thonburi (KMUTT)}\\
Bangkok, Thailand \\
jonathan@sit.kmutt.ac.th}}

\maketitle

\begin{abstract}
Freezing of gait is a Parkinson’s Disease symptom that episodically inflicts a patient with the inability to step or turn while walking. While medical experts have discovered various triggers and alleviating actions for freezing of gait, the underlying causes and prediction models are still being explored today. Current freezing of gait prediction models that utilize machine learning achieve high sensitivity and specificity in freezing of gait predictions based on time-series data; however, these models lack specifications on the type of freezing of gait events. We develop various deep learning models using the transformer encoder architecture plus Bidirectional LSTM layers and different feature sets to predict the three different types of freezing of gait events. The best performing model achieves a score of 0.427 on testing data, which would rank top 5 in Kaggle’s Freezing of Gait prediction competition, hosted by THE MICHAEL J. FOX FOUNDATION. However, we also recognize overfitting in training data that could be potentially improved through pseudo labelling on additional data and model architecture simplification.
\end{abstract}

\begin{IEEEkeywords}
Parkinson’s Disease, Freezing of Gait, Machine Learning
\end{IEEEkeywords}

\section{Introduction}
Freezing of gait (FOG) is a common Parkinson’s disease (PD) mobility disturbance that episodically inflicts PD patients with the inability to step or turn while walking. In advancing stages of PD, 60\% of PD patients could experience FOG events [1]; each FOG event could last up to a few minutes. FOG episodes often occur at the initialization of walking (start hesitation), turning, or during walking periods, during which PD patients would experience dystonic gait during the “on” state and hypokinetic gait during the “off” state of FOG [2].

While medical experts can recognize conditions that precipitate to FOG, such as narrow passages, being time pressure, distractions, dual-tasking, and male sex [3, 4] and actions that could alleviate FOG, such as emotion, excitement, and auditory cueing [3], they are still trying to find underlying causes and develop reasonable prediction models for FOG events. Various machine learning and data analysis techniques have been used to detect FOG events and pre-FOG events from time-series data.

One FOG detection algorithm uses pure statistical inference on lower-limb acceleration to detect FOG events [5]. Vertical, horizontal forward, and horizontal lateral acceleration data are collected from sensors located at PD patients’ shank, thigh, and lower back as they perform three walking tasks. Experiments are videotaped to allow medical experts to annotate onsets and durations of FOG episodes. This algorithm predominantly relies on three features – Freeze Index (FI) and Freezing Index threshold, Wavelet Mean (WM), and Sample Entropy (SE) – extracted on acceleration data over sliding windows of 2 to 4 second durations [5]. Naghavi et al. discovered that using the [Sensor at Thigh, vertical acceleration, 4-second sliding window, FI] and the [Sensor at Shank, vertical acceleration, 3-second sliding window, FI] produces the highest predictivity of $96.7\%\pm5.6\%$ and $95.7\%\pm3.9\%$, respectively. However, while using patient-dependent FI thresholds achieves around 90\% FOG detection [5, 6], using a global FI threshold of 11 subjects achieves only 78\% FOG detection [6]. Due to the considerable post-processing time required in generating optimal FI thresholds for different patients, the FI algorithm is not a viable method to generalize FOG detection for larger groups of PD patients.

One machine learning model uses time-series plantar pressure data from 11 PD patients to detect FOG events. Each PD patient is required to complete a 25-meter walking task, during which a set of 16 features related to the center of pressure coordinates, center of pressure velocities, center of pressure accelerations, and ground reaction forces is collected from plantar pressure sensors. A 2-layer LSTM neural network architecture and a 3-layer LSTM neural network architecture show similar performance, achieving 82.1\% mean sensitivity and 89.5\% mean specificity and 83.4\% mean sensitivity and 87.7\% mean specificity in leave-one-out cross validation detection, respectively. However, plantar pressure insole sensors in the research are for single use, which means that this detection system cannot generalize to larger scale experiments or real-life detection systems [1].

Another model collects acceleration data from inertial sensors on the shins of 11 PD patients as they perform a 7-meter TUG test. Feature extraction includes standard deviation, angular jerk, power spectral entropy, principal harmonic frequency, etc. The experiment uses a decision tree for feature selection and SVM for FOG/non-FOG classification. It achieves 85.5\% sensitivity and 86.3\% specificity for pre-FOG detection and 88.0\% sensitivity and 90.3\% specificity for FOG episode detections in leave-one-out cross validation [7]. However, this model also does not detect the type of FOG onset, which may be important to researcher’s understanding of FOG occurrences as well.

In Feature extraction for robust physical activity recognition, Zhu et al. mentions various feature extractions from body motion acceleration data that could be used for physical activity classification. Features used in the time domain includes original accelerometer signals, magnitude of combined signals, jerk signals, magnitude of combined jerk signals, while features from the frequency domain includes Fast Fourier Transforms (FFT) from original signals, FFT from magnitude signal, and FFT from jerk signal. Further feature extractions include mean value, standard deviation, signal magnitude area, skewness, kurtosis, etc. Results from the experiment reveal that time-domain feature models consistently outperform frequency-domain feature models and time-frequency-combination models [8]. Therefore, feature selections in this paper primarily focus on manipulating acceleration data in the time-domain.

In accordance with Kaggle’s Parkinson’s Freezing of Gait Prediction competition, this paper aims to discover the most suitable deep learning model that can detect FOG episodes and specify the type of FOG onset (start hesitation, turn, or walking) from time-series acceleration data. Model development draws on data-preprocessing, feature extraction, model architecture, and training techniques utilized in related prior works, with a particular emphasis on using different feature sets to determine which features produce more robust models.

\section{Method}
\subsection{Data Input}
Defog data files, Tdcsfog data files, Notype data files, Defog metadata, Tdcsfog metadata, and subject data file generously provided by THE MICHAEL J. FOX FOUNDATION are used in model training. Defog, Tdcsfog, and Notype data files are time-series acceleration data collected from a wearable 3D acceleration sensor on Parkinson’s Disease patients’ lower back; Defog metadata and Tdcsfog metadata contain information about each time-series data file; subject data file contains information about each testing participant. [9]
\vspace{0.5cm}

Defog acceleration data are collected in Parkinson’s Disease patients’ home as they perform a FOG-provoking test protocols. There are 91 Defog files with an average of 148634 timeframes per file. Each data file consists of vertical acceleration (AccV), mediolateral acceleration (AccML), and anteroposterior acceleration (AccAP) values in units of $9.81\frac{m}{s^2}$ collected at 124 Hz. Each trial is also recorded on camera; medical experts review each video tape and annotate time steps where the patient experiences start hesitation, turn, or walking FOG onset. The filename indicates the trial ID (ex: 02ab235146.csv). [9]
\begin{figure}[htbp]
\centerline{\includegraphics{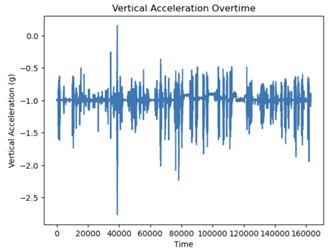}}
\caption{Sample vertical acceleration data from Defog Dataset}
\label{fig}
\end{figure}

Notype acceleration data come from the Defog dataset. There are 46 Notype files with an average of 222850 timeframes per file. However, FOG occurrence annotation lacks type specification. Usage of Notype data files in pseudo labelling is documented in Section 4.4 Pseudo Labelling. [9]

Tdcsfog acceleration data are collected in a medical lab as they perform a FOG-provoking test protocols. There are 833 Tdcsfog files with an average of 8479 timeframes per file. Each data file consists of AccV, AccML, and AccAP values in units of $\frac{m}{s^2}$ collected at 100 Hz. Each trial is also recorded on camera; medical experts review each video tape and annotate time steps where the patient experiences start hesitation, turn, or walking FOG occurrence. The filename indicates the trial ID (ex: 003f119e14.csv). [9]
\vspace{0.5cm}

Defog Metadata and Tdcsfog Metadata files consists of time trial IDs, corresponding subject IDs, and the medication status of the patient (“on” or “off”) at the time of walking. [9]
\vspace{0.5cm}

Subject data file consists of each patient’s age, years since Parkinson’s Disease diagnosis, UPDRS score during medication “on” state, UPDRS score during medication “off” state, and NFOGQ score [9]. UPDRS score is a measurement of a PD patient’s overall PD condition determined by four components: mentation, behaviour, and mood (Part I), activities and daily living (Part II), motor symptoms (Part III), and complications of therapy (Part IV), where a higher score indicates more severe PD conditions [10]. NFOGQ score is a score that PD patients give themselves based on their personal assessment of FOG condition. The minimum scores of zero are given by people who do not have FOG [9].

\subsection{Model Construction}
A total of 14 preliminary model groups are created based on different combinations of data inputs, feature sets, and the same architecture. Each model group consists of three models generated from three iterations of three-fold cross validations. Each round of model group testing involves taking the average of three predictions generated by the three models in that model group.
Six Defog model groups are subsequently used to generate “semi-pseudo-labelled” data with Notype training data and retrained with additional data to create six additional model groups. 
\vspace{0.5cm}

Preliminary data analysis on Defog time-series data files and Tdcsfog time-series data files reveals a notable separation between these two datasets (Fig. 2). Therefore, each feature set model group is separated into a Defog model group and a Tdcsfog model group, denoted by DEFOG \#X and TDCSFOG \#X. One competition submission combines DEFOG \#X and TDCSFOG \#X to make predictions.

The mean, maximum, minimum, and standard deviation of each acceleration type in all time-series data files are collected as feature values, forming a 12-dimensional feature vector for each of the 924 time-series data files. The $924\times12$ data matrix undergoes dimension reduction to a $924\times2$ data matrix with Scikit-Learn’s Principal Component Analysis method. The two features of each data point are plotted against each other on a 2D graph (Fig. 2). Clustering data points by data type (Defog or Tdcsfog) results in a silhouette score of 0.906.
\begin{figure}[htbp]
\centerline{\includegraphics{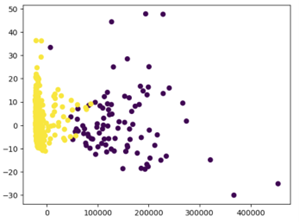}}
\caption{Clustering of Tdcsfog data (yellow) and Defog data (purple) after dimension reduction with PCA.}
\label{fig}
\end{figure}

The TransBiLSTM model is inspired by the first-place winner solution of the Parkinson’s Disease Freezing of Gait Prediction competition, which uses 5 transformer encoder layers, 3 Bidirectional LSTM layers, and one final Dense layer that outputs the probability of each type of FOG event (Fig. 3) [11]. The encoder layer consists of one multi-head attention layer with 6 attention heads of size 320, one layer normalization layer, and one fully connected sequential block with two dense layers (320) and two dropout layers (0.1). The BiLSTM layers produce 320-dimensional outputs.
\begin{figure}[htbp]
\centerline{\includegraphics{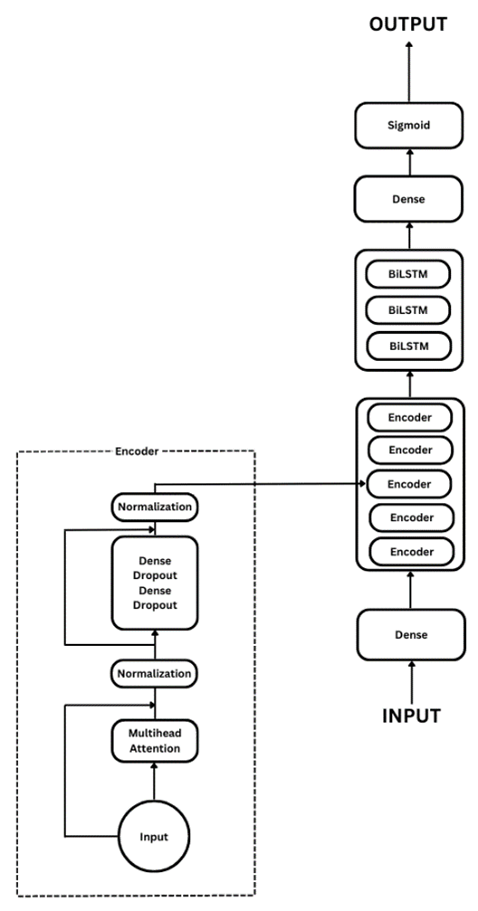}}
\caption{Transformer Encoder + BiLSTM model architecture}
\label{fig}
\end{figure}
The purpose of feature set generation is to develop various feature combinations and discover the most appropriate feature extractions for this set of FOG data.

\subsubsection{Feature Set A}
AccV, AccML, AccAP - These are the “baseline” features.
\subsubsection{Feature Set B}
AccV, AccML, AccAP, TimeFrac - Feature Set A plus normalized time.
\subsubsection{Feature Set C}
AccV, AccML, AccAP, TimeFrac, JerkV, JerkML, JerkAP - Feature Set B plus the jerk of acceleration features.

\subsubsection{Feature Set D}
AccV, AccML, AccAP, TimeFrac, JerkV, JerkML, JerkAP, AccM, JerkM - Feature Set C plus the acceleration magnitude and its jerk.

\subsubsection{Feature Set E}
AccV, AccML, AccAP, TimeFrac, JerkV, JerkML, JerkAP, Gender, Medication - Feature Set C plus Gender and Medication information from metadata file.

\subsubsection{Feature Set F}
AccV, AccML, AccAP, TimeFrac, JerkV, JerkML, JerkAP, AccM, JerkM, Gender, Medication - Feature Set D plus Gender and Medication information from metadata file.

\subsubsection{Feature Set G}
AccV, AccML, AccAP, TimeFrac, JerkV, JerkML, JerkAP, Subject Cluster - Feature Set C plus clustering of subjects.
\vspace{0.5cm}

"Semi pseudo labelling" is performed on Notype data files, as inspired by [12]. First, one of the four highest-scoring preliminary Defog model groups, DEFOG MODEL \#X, is chosen to predict FOG probability for each class for each timeframe of a Notype time-series file using acceleration data and any additional features used in that preliminary model. Then, Start Hesitation, Turn, and Walking columns are created, where:

\begin{itemize}
\item Start Hesitation is one in rows where Event=1 and Start Hesitation probability is highest.
\item Turn is one in rows where Event=1 and Turn probability is highest.
\item Walking is one in rows where Event=1 and Walking probability is highest.
\end{itemize}
\vspace{0.5cm}

Two important metrics are used to monitor the performance of model training: Mean Absolute Error (MAE) and Mean Average Precision Score (MAP).

During each fold of model training, the performance of a model is monitored by the validation mean absolute error; the training epoch that produces a model with the lowest validation mean absolute error is saved to the output. After each fold of cross validation, the performance of the model is determined by the MAP of its predictions on the out-of-fold validation data and the true labels of the out-of-fold validation data. Finally, the performance of a model group is determined by the average of the MAP’s produced by models in that model group.

The performance of a feature set, which consists of a Defog model group and a Tdcsfog model group, is determined by the MAP of each model group and the proportions of Defog data and Tdcsfog data. Namely,

\begin{center}
\centering
Feature Set Performance (FP)

=

$0.657\cdot DMAP + 0.343\cdot TMAP$
\end{center}

Where 0.657 is the proportion of Defog data, DMAP is the mean average precision of the Defog model, 0.343 is the proportion of Tdcsfog data, and TMAP is the mean average precision of the Tdcsfog model.

\section{Results}
Each model group is trained on a 0.67/0.33 training and validation data split. Testing data consists of about 250 time-series data files consisting of Defog and Tdcsfog data in similar proportion to the training dataset. Feature Set C model groups produced the best predictions with a combined score of 0.427 based on the private MAP score (68\% of testing data) and the public MAP score (32\% of testing data) generated by Parkinson’s FOG Prediction Kaggle Competition.

\subsection{Feature Set Performance for Preliminary Model Groups}
The Defog model group and The Tdcsfog model group for each feature set are combined into one prediction instance on Defog data and Tdcsfog data, respectively. The DMAP, TMAP, FP, and public score, private score, and combined score for each prediction submission are listed in Table 1.

\begin{table}[htbp]
\begin{center}
\caption{Model MAP and submission scores for each feature set (Preliminary).}
\centering
\begin{tabularx}{\columnwidth}{|X|X|X|X|X|X|X|}
\hline
Feat. Set & DMAP & TMAP & FP & Private Score & Public Score & Total Score  \\
\hline
A & 0.224 &	0.642 & 0.367 &	0.330 & 0.323 & 0.328 \\
\hline
B & 0.250 &	0.687 & 0.400 & 0.362 & 0.374 & 0.366 \\
\hline
C & 0.214 &	0.669 & 0.370 &	0.420 & 0.393 &	\textbf{0.411} \\
\hline
D & 0.237 &	0.686 & 0.391 &	0.342 & 0.372 &	0.352 \\
\hline
E & 0.239 &	0.659 & 0.383 &	0.400 & 0.342 &	0.381 \\
\hline
F & 0.227 &	0.652 & 0.373 &	0.391 & 0.352 &	0.378 \\
\hline
G & 0.112 &	0.601 & 0.280 &	0.156 & 0.238 &	0.183 \\
\hline
\end{tabularx}
\end{center}
\end{table}

\subsection{Effect of Pseudo Labelling}
The DMAP, TMAP, FP, and public score, private score, and combined score for each prediction submission are listed in Table 2.

\begin{table}[htbp]
\begin{center}
\caption{Model MAP and submission scores for each feature set (Improved).}
\centering
\begin{tabularx}{\columnwidth}{|X|X|X|X|X|X|X|}
\hline
Feat. Set & DMAP & TMAP & FP & Private Score & Public Score & Total Score  \\
\hline
A &	0.300 &	0.642 &	0.417 &	0.356 &	0.328 &	0.347\\
\hline
B &	0.251 &	0.687 &	0.400 &	0.377 &	0.376 &	0.377\\
\hline
C &	0.308 &	0.669 &	0.432 &	0.443 &	0.392 &	\textbf{0.427}\\
\hline
D &	0.264 &	0.686 &	0.409 &	0.349 &	0.374 &	0.357\\
\hline
E &	0.310 &	0.659 &	0.430 &	0.408 &	0.350 &	0.389\\
\hline
F &	0.262 &	0.652 &	0.395 &	0.397 &	0.357 &	0.384\\
\hline

\hline
\end{tabularx}
\end{center}
\end{table}

New Defog Model groups trained on pseudo labelled data allow feature set performances to increase by an average of 9\% and testing MAP scores to increase by an average of 3\%. Feature Set C and Feature Set E have the highest FP of 0.432 and 0.430, respectively. These two feature sets also have the highest submission score, where Feature Set C produced a score of 0.427 and Feature Set E produced a score of 0.389. A score of 0.427 ranks top 5 in Kaggle's Freezing of Gait Competition, where the top 5 scores are 0.514, 0.451, 0.436, 0.417, and 0.390 [9].

\section{Discussions}

Preliminary model group submissions reveal a minimal correlation between MAP scores obtained in training and testing MAP scores. There is a -0.305 correlation coefficient between Defog model MAP scores and testing MAP scores, a 0.191 correlation between Tdcsfog model MAP scores and testing MAP scores, and a -0.106 overall correlation coefficient between FP and testing MAP scores. These results show that model groups likely suffered from overfitting in the training processes.

Retraining Defog models with semi pseudo labelled data significantly increases the correlation between MAP scores obtained in training and testing MAP scores. In particular, the correlation coefficient between Defog model MAP scores and testing MAP scores increased to 0.354, which prompts the correlation coefficient between FP and testing MAP scores to 0.455. Moreover, the average performance of the six new competition submissions increased by 3\% compared to preliminary model submissions, with the best performance increasing by 3.7\%. These results suggest that training with additional data can result in more robust and better-performing models.

The significance of Feature Set G is the addition of subject clustering information. Reduction in prediction performance compared to its basis feature set – Feature Set C – suggests that patients’ Parkinson’s Disease statuses are excessive information in a real-time FOG prediction system based primarily on motion sensor data.

On average, models that use jerk perform 7.5\% better in testing MAP scores than models that do not use jerk as features. Despite the overfitting in models, this result is consistent with FP scores as well, where models that use jerk features perform 2\% better in FP scores. The importance of jerk corresponds with the fact that FOG is an event of sudden muscle contraction that may cause a person to experience sudden change in body acceleration.

On top of the low correlation between validation MAP scores and testing MAP scores, overfitting also causes different models to produce low-variance testing MAP scores despite different feature additions. The primary reason for causing significant overfitting in these FOG prediction models may be the overly complicated neural network created during model training. Using different feature sets in model training produces models that have a minimum of 20,180,483 parameters to a maximum of 20,218,883 parameters. However, while the enormous network specializes in the training data, they do not generalize well to unseen data, and this complex structure reduces the significance of the new information brought by new features. As a result, it is hard to accurately determine the effect that each feature has on model performance.

\section{Conclusion}
The best performing model that uses time, acceleration signals, and jerks of acceleration and the transformer encoder plus bidirectional LSTM layers architecture achieves a 0.427 MAP, which would rank top 5 in Kaggle’s Parkinson’s Freezing of Gait competition [9]. The overall performance improvement with adding jerk as features imply that the change of acceleration features can positively impact FOG prediction. However, we recognize that there is overfitting to training data that causes noticeable change between training validation MAP scores and testing MAP scores. The robustness of models considerably improved after one round of semi pseudo labelling, which implies that further pseudo labelling on a greater variety of data can potentially alleviate the severity of overfitting. We would also like to see in the future if simplifying the model architecture and using more regularization techniques can also improve the robustness of these models. Furthermore, we recognize that this model does not provide real-time FOG predictions; nevertheless, we hope that results from our machine learning approach can inspire better FOG event-specific prediction models in the future.

\section*{Acknowledgment}

The authors would like to thank all participants in the Parkinson’s Freezing of Gait Competition for providing help along the way.

\end{document}